\documentclass[preprint,12pt]{elsarticle}

\usepackage{amsmath,amssymb}
\usepackage{booktabs}
\usepackage{graphicx}
\usepackage{xcolor}
\usepackage{hyperref}
\usepackage{multirow}
\usepackage{subcaption}
\usepackage{array}
\usepackage{float}
\usepackage{placeins}
\graphicspath{{figures/}}

\hypersetup{colorlinks=true,linkcolor=blue,citecolor=blue,urlcolor=blue}
\journal{The Journal of Supercomputing}

\begin{document}

\begin{frontmatter}

\title{Component Ablation for Efficient Hybrid Language Model Architectures: Performance, Resilience, and Compression Implications}

\author[inst1]{Hector Borobia\fnref{cor1}}
\ead{hecboar@doctor.upv.es}
\author[inst2]{Elies Segu\'i-Mas}
\ead{eseguim@upvnet.upv.es}
\author[inst3]{Guillermina Tormo-Carb\'o}
\ead{gtormo@omp.upv.es}
\fntext[cor1]{Corresponding author.}
\affiliation[inst1]{organization={VRAIN -- Valencian Research Institute for Artificial Intelligence, Universitat Polit\`ecnica de Val\`encia},
    city={Valencia},
    country={Spain}}
\affiliation[inst2]{organization={Department of Economics and Social Sciences, Universitat Polit\`ecnica de Val\`encia},
    city={Valencia},
    country={Spain}}
\affiliation[inst3]{organization={Department of Business Organisation, Universitat Polit\`ecnica de Val\`encia},
    city={Valencia},
    country={Spain}}

\begin{abstract}
Hybrid language models combine softmax attention with linear-time sequence mechanisms to improve inference efficiency, but the operational role of each component remains under-characterized. We analyze two sub-1B hybrid architectures---Qwen3.5-0.8B, with sequential Gated DeltaNet and attention layers, and Falcon-H1-0.5B, with parallel Mamba-2 and attention paths---using Qwen2.5-0.5B as a Transformer control. We apply reversible group, layer-wise, positional, and matched random ablations, evaluating five downstream benchmarks, WikiText-2 perplexity, and hidden-state diagnostics. Both hybrid components materially affect performance, but likelihood is more sensitive to removing the linear/SSM pathway than to removing attention. Ablation impact is strongest in early or mid-network components and varies across task families. Matched random controls further show lower normalized perplexity degradation in the hybrid model than in the Transformer control. These results support component-aware analysis for hybrid-model compression, architecture design, and robust deployment.
\end{abstract}

\begin{keyword}
hybrid language models \sep state space models \sep linear attention \sep architecture analysis \sep model compression \sep performance degradation \sep resilient inference
\end{keyword}

\end{frontmatter}

\FloatBarrier
\section{Introduction}
\label{sec:introduction}

The Transformer architecture~\cite{vaswani2017attention} has been the dominant paradigm for large language models (LLMs), with softmax self-attention serving as its core sequence-mixing mechanism. However, the quadratic computational complexity of attention with respect to sequence length has motivated the development of alternative approaches. State space models (SSMs)~\cite{gu2022efficiently}, particularly their selective variants such as Mamba~\cite{gu2023mamba} and Mamba-2~\cite{dao2024transformers}, offer linear-time alternatives with competitive performance. Similarly, linear attention mechanisms based on gated recurrences~\cite{yang2024gated} provide efficient alternatives to softmax attention while maintaining expressiveness through data-dependent gating.

Rather than choosing between attention and these alternatives, recent work has converged on \textit{hybrid architectures} that combine both. The Falcon-H1 family~\cite{zuo2025falcon} employs a \textit{parallel hybrid} design where each block processes its input through both a Mamba-2 SSM and an attention head simultaneously, summing their outputs. Qwen3.5~\cite{qwen2025qwen3} adopts a \textit{sequential hybrid} approach, interleaving dedicated linear attention layers (based on Gated DeltaNet) with full softmax attention layers throughout the network depth.

A critical question arises: \textbf{do hybrid models actually use both components, or does one pathway dominate while the other is effectively bypassed?} Recent work by Benfeghoul et al.~\cite{benfeghoul2025untangling} found that in post-training linearization methods, the linear attention component is often inadvertently bypassed. Wang et al.~\cite{wang2025systematic} showed that the choice of hybridization ratio significantly affects recall performance, suggesting that the interplay between components is non-trivial.

Despite the proliferation of hybrid architectures, systematic empirical studies of how each component contributes to model behavior remain scarce. Most architectural papers report aggregate benchmarks but do not decompose performance by component. This leaves practitioners without guidance on which components to preserve during compression, which layers to prioritize during distillation, or how to design more efficient hybrid configurations.

From a high-performance deployment perspective, hybrid language models raise a practical systems question: which architectural components must be preserved to maintain graceful degradation under compression, pruning, or partial structural failure? Standard aggregate benchmarks do not answer this question because they do not isolate the contribution of attention, SSM, or linear-attention pathways. We therefore study component-level performance degradation under reversible ablation, using matched random controls and a Transformer baseline to separate component identity from generic layer-removal effects. This framing connects hybrid architecture analysis to efficient inference, model compression, and robust deployment.

\paragraph{Contributions} This paper presents a systematic functional ablation analysis of two representative hybrid language model designs: a sequential Gated DeltaNet/attention architecture and a parallel Mamba-2/attention architecture. The main contributions are:

\begin{enumerate}
    \item We define a reversible component-ablation protocol for hybrid language models, covering group-level, layer-wise, positional, and matched random ablations under a unified evaluation procedure.

    \item We show that, in the two tested natively trained hybrids, removing either attention or the alternative sequence-processing pathway substantially degrades downstream performance, indicating measurable functional use of both component types.

    \item We find that WikiText-2 perplexity is more sensitive to removing the linear-attention/SSM pathway than to removing attention, suggesting that these pathways dominate next-token likelihood under the tested ablation interventions.

    \item We quantify a positional sensitivity pattern in which many of the strongest performance drops are concentrated in early or mid-network components, providing empirical constraints for position-aware pruning and distillation strategies.

    \item We compare targeted ablations with matched random controls and a same-family Transformer control, showing that component identity, layer position, and architecture type all affect degradation under structural removal.
\end{enumerate}

\FloatBarrier
\section{Related Work}
\label{sec:related}

\paragraph{Efficient sequence modeling and hybrid language architectures}
The idea of combining attention with recurrent or state-space mechanisms has gained traction as a solution to the efficiency--expressiveness trade-off. Selective state space models such as Mamba~\cite{gu2023mamba} and Mamba-2~\cite{dao2024transformers} provide linear-time sequence processing, while gated linear attention mechanisms~\cite{yang2024gated} offer another route to efficient long-context modeling. Recent hybrid systems combine these mechanisms with softmax attention. Samba~\cite{ren2024samba} demonstrated that a hybrid of Mamba and sliding-window attention achieves strong performance with unlimited context, including ablation studies across model sizes up to 1.7B. Taipan~\cite{nguyen2024taipan} combines Mamba-2 with selective attention layers that dynamically identify tokens requiring long-range interactions. Kimi Linear~\cite{kimi2025linear} extends Gated DeltaNet with a finer-grained gating mechanism and reports reduced KV-cache requirements.

\paragraph{Hybridization ratios and architecture design}
The systematic analysis by Wang et al.~\cite{wang2025systematic} trained 72 models at 340M and 1.3B parameters, evaluating six linear-attention variants across five hybridization ratios. They found that superior standalone linear models do not necessarily excel in hybrid configurations, and that a linear-to-full-attention ratio between 3:1 and 6:1 achieves Transformer-level recall efficiently. Ren et al.~\cite{ren2024exploring} analyzed the limitations of Mamba in copy and chain-of-thought reasoning tasks, showing that while constant-sized Mamba struggles with certain operations, hybridization with attention can overcome these limitations. These studies motivate component-level analysis: if the performance of a hybrid depends on component ratio and placement, then compression and architecture design require knowing which component types and positions are most sensitive.

\paragraph{Component analysis and ablation methodology}
Ablation has a long history as a diagnostic method for neural networks~\cite{meyes2019ablation}. In language models, mechanistic interpretability has studied how individual components contribute to model behavior, including work on attention circuits~\cite{elhage2021mathematical,olsson2022context}. Those studies focus mainly on Transformer architectures. For hybrid models, Benfeghoul et al.~\cite{benfeghoul2025untangling} identified that some post-training linearized hybrids inadvertently bypass their linear component, relying primarily on sliding-window attention. Our work differs by studying \textit{natively trained} hybrids rather than post-hoc conversions. Haller et al.~\cite{haller2025matters} studied what factors matter in linearizing language models, providing background on why gated delta-rule variants behave differently from additive linearized models.

\paragraph{Compression, robustness, and deployment motivation}
Hybrid models are often motivated by efficient inference, memory reduction, and long-context deployment. For such settings, aggregate benchmark scores are insufficient: a compressed or partially degraded model may retain similar average accuracy while relying on different components or suffering sharply on likelihood metrics. Component ablation therefore provides a practical tool for identifying which mechanisms are risky to prune, which layers require higher fidelity, and whether a hybrid architecture degrades more gracefully under structural perturbation. This connects the present study to architecture-level performance analysis and robust deployment of efficient neural models.

\FloatBarrier
\section{Experimental Setup}
\label{sec:setup}

\FloatBarrier
\subsection{Models}

We study two architecturally distinct hybrid language models in their base (pre-trained, not instruction-tuned) variants:

\paragraph{Qwen3.5-0.8B}~\cite{qwen2025qwen3} is a 752M-parameter model with 24 decoder layers in a \textit{sequential hybrid} pattern. It interleaves linear attention layers based on Gated DeltaNet~\cite{yang2024gated} with full softmax attention layers. Of the 24 layers, 18 use linear attention and 6 use full attention (3:1 ratio). Each layer contains only one mechanism type. The model supports a maximum context of 262,144 tokens.

\paragraph{Falcon-H1-0.5B}~\cite{zuo2025falcon} is a 521M-parameter model with 36 decoder blocks in a \textit{parallel hybrid} design. Each block contains both a Mamba-2 SSM path~\cite{dao2024transformers} and an attention path that process the input simultaneously. The outputs of both paths are combined within each block, meaning every layer contains both component types. It supports a context length of 16,384 tokens.

The choice of sub-1B models is deliberate: it enables exhaustive layer-wise ablation (84 conditions for Falcon, including individual ablation of each component in each of 36 layers) while keeping the analysis reproducible on a single GPU. The architectural contrast---sequential vs.\ parallel hybridization---is the primary variable of interest. As an additional control, we include \textbf{Qwen2.5-0.5B}~\cite{qwen2025qwen25}, a pure Transformer (490M parameters, 24 layers, softmax attention only) from the same model family, to determine whether observed patterns are hybrid-specific or general properties of deep networks.

\paragraph{Scope of model selection} We focus on sub-1B checkpoints because they make exhaustive component-level sweeps feasible under a reproducible single-GPU protocol. The goal is not to claim scale-invariant behavior across all hybrid LLMs, but to isolate functional differences between two representative hybridization strategies---sequential and parallel---under controlled ablation. We therefore interpret the results as architecture-level evidence for the tested model families, and as hypotheses for larger-scale validation rather than as universal claims about all hybrid language models.

\FloatBarrier
\subsection{Ablation Mechanism}
\label{sec:ablation_mechanism}

We implement reversible ablation through two strategies adapted to each architecture:

\paragraph{Sequential skip (Qwen)} For models with distinct layer types, we replace the layer's output with its input, effectively converting the layer to an identity function within the residual stream: $\mathbf{h}_{l+1} = \mathbf{h}_l$ instead of $\mathbf{h}_{l+1} = \mathbf{h}_l + f_l(\mathbf{h}_l)$.

\paragraph{Parallel zeroing (Falcon)} For models with parallel components within each block, we zero out the output contribution of the targeted component via forward hooks: if the original block computes $\mathbf{h}_{l+1} = \mathbf{h}_l + f_{\text{attn}}(\mathbf{h}_l) + f_{\text{ssm}}(\mathbf{h}_l)$, then ablating SSM yields $\mathbf{h}_{l+1} = \mathbf{h}_l + f_{\text{attn}}(\mathbf{h}_l)$.

Both mechanisms are implemented as context-managed forward hooks that are cleanly reversible, preserving model weights for subsequent conditions.

\paragraph{Intervention granularity} The intervention necessarily follows the computational structure exposed by each architecture. In Qwen, disabling a sequential hybrid layer replaces the full layer transformation with an identity map, thereby removing the token mixer and the layer-local feed-forward transformation. In Falcon, disabling a path zeros only the selected parallel component while preserving the other path and the block-level feed-forward computation. Consequently, cross-model ablation magnitudes should not be compared as identical interventions. Our main comparisons are within-model contrasts, calibrated by matched random controls.

\FloatBarrier
\subsection{Evaluation Protocol}

\paragraph{Benchmarks} We evaluate on five standard benchmarks covering diverse capabilities: MMLU~\cite{hendrycks2021measuring} (5-shot, $n$=2,000; knowledge), GSM8K~\cite{cobbe2021training} (8-shot, $n$=50--100; mathematical reasoning), ARC-Challenge~\cite{clark2018think} (25-shot, $n$=299; science reasoning), HellaSwag~\cite{zellers2019hellaswag} (10-shot, $n$=2,000; commonsense), and TruthfulQA-MC~\cite{lin2022truthfulqa} (0-shot, $n$=790; factuality). For GSM8K, baseline and group ablation conditions use $n$=100, while positional ablations use $n$=50 due to compute constraints.

\paragraph{Benchmark sample sizes} The benchmark subsets are intended to provide a consistent diagnostic signal across many ablation conditions rather than definitive leaderboard estimates. GSM8K uses a smaller subset in positional ablations because these sweeps multiply the number of evaluated conditions. We therefore interpret GSM8K results primarily by direction and relative degradation pattern. All comparisons involving different sample sizes are interpreted as diagnostic evidence of relative sensitivity rather than final task-performance estimates.

\paragraph{Ablation conditions} We evaluate four categories totaling 84+ conditions per model:
(i)~\textit{Group ablations}: removing all instances of one component type simultaneously;
(ii)~\textit{Layer-wise ablations}: removing a single component in a single layer, sweeping across all layers;
(iii)~\textit{Positional ablations}: removing all instances of a component within early, middle, or late thirds of the network (Falcon);
(iv)~\textit{Matched random controls}: removing the same number of randomly selected computational units. For sequential models, the control removes the same number of layers as the targeted condition. For parallel models, the control samples the same number of component paths from the set of available attention and SSM paths. Reported random-control values are averages over trials and should be interpreted together with their variance.

\paragraph{Statistical validation} For all group ablation comparisons, we compute bootstrap confidence intervals (1,000 resamples) on the score drop relative to baseline. A confidence interval excluding zero indicates statistical significance at the 95\% level.

\FloatBarrier
\section{Results}
\label{sec:results}

\FloatBarrier
\subsection{Group-Level Performance Degradation Under Component Removal}
\label{sec:exp1}

Table~\ref{tab:group_ablation} presents the group ablation results across all five benchmarks for both models. Removing either component type substantially degrades performance in both architectures, indicating measurable functional use of both component types under the tested interventions.

\begin{table}[!htbp]
\centering
\caption{Group ablation results across five benchmarks. Values show absolute score; parentheses show change from baseline ($\Delta$). Conditions marked with $\ast$ indicate that the mean score drop across all benchmarks has a bootstrap 95\% CI excluding zero. Note that individual benchmarks may vary: TruthfulQA shows non-significant drops under some conditions.}
\label{tab:group_ablation}
\small
\resizebox{\textwidth}{!}{%
\begin{tabular}{@{}llccccc@{}}
\toprule
\textbf{Model} & \textbf{Condition} & \textbf{MMLU} & \textbf{GSM8K} & \textbf{ARC-C} & \textbf{HSwag} & \textbf{TFQA} \\
\midrule
\multirow{3}{*}{\shortstack[l]{Qwen3.5\\0.8B}}
  & Baseline             & .489  & .350  & .742  & .516  & .285  \\
  & Linear off$^\ast$    & .250\,{\scriptsize(-.239)} & .000\,{\scriptsize(-.350)} & .261\,{\scriptsize(-.482)} & .268\,{\scriptsize(-.248)} & .230\,{\scriptsize(-.054)} \\
  & Attention off$^\ast$ & .214\,{\scriptsize(-.274)} & .000\,{\scriptsize(-.350)} & .227\,{\scriptsize(-.515)} & .341\,{\scriptsize(-.175)} & .267\,{\scriptsize(-.018)} \\
\midrule
\multirow{3}{*}{\shortstack[l]{Falcon-H1\\0.5B}}
  & Baseline             & .568  & .400  & .736  & .512  & .223  \\
  & SSM off$^\ast$       & .224\,{\scriptsize(-.344)} & .030\,{\scriptsize(-.370)} & .211\,{\scriptsize(-.525)} & .265\,{\scriptsize(-.247)} & .263\,{\scriptsize(+.041)} \\
  & Attention off$^\ast$ & .228\,{\scriptsize(-.340)} & .000\,{\scriptsize(-.400)} & .214\,{\scriptsize(-.522)} & .369\,{\scriptsize(-.143)} & .227\,{\scriptsize(+.004)} \\
\bottomrule
\end{tabular}}%
\end{table}

For Qwen, removing all 18 linear attention layers reduces MMLU from .489 to .250 (near chance for a 4-way task) and GSM8K from .350 to .000 (score falling to 0 on the evaluated GSM8K sample). Removing the 6 full attention layers causes comparable MMLU degradation (.214) and identical GSM8K collapse. Notably, removing attention has a \textit{smaller} effect on HellaSwag (.341 vs.\ .268), suggesting that linear attention contributes more to commonsense reasoning.

\begin{figure}[!htbp]
    \centering
    \includegraphics[width=\textwidth]{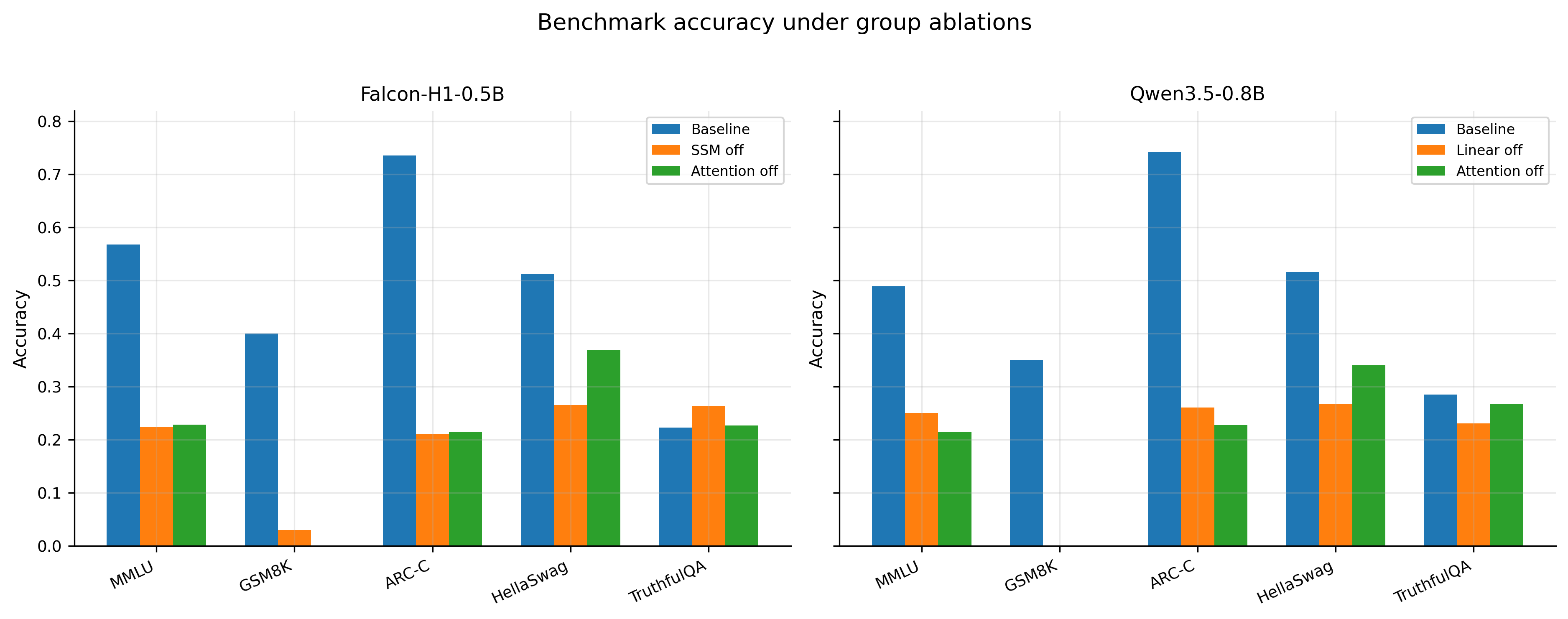}
    \caption{Benchmark accuracy under baseline and group-level component ablations for Qwen3.5-0.8B and Falcon-H1-0.5B. Values correspond to the scores in Table~\ref{tab:group_ablation}. Lower values indicate greater performance degradation after removing the specified component type. The plot summarizes task-level effects and should be interpreted together with the per-benchmark deltas in Table~\ref{tab:group_ablation}.}
    \label{fig:radar}
\end{figure}

For Falcon, both SSM-off and attention-off yield nearly identical MMLU (.224 vs.\ .228) and ARC-C (.211 vs.\ .214) degradation. The main asymmetry is in HellaSwag, where SSM removal causes greater damage (.265) than attention removal (.369). Figure~\ref{fig:radar} shows these profiles.

\FloatBarrier
\subsection{Perplexity Reveals Asymmetric Likelihood Sensitivity}
\label{sec:perplexity}

While benchmark accuracy shows similar degradation magnitudes for both components, perplexity analysis reveals a strong asymmetry in likelihood sensitivity (Table~\ref{tab:perplexity}).

\begin{table}[!htbp]
\centering
\caption{WikiText-2 perplexity under group ablation. Removing the linear-attention pathway in Qwen3.5-0.8B and the SSM pathway in Falcon-H1-0.5B produces larger likelihood degradation than removing attention in the corresponding model. Multipliers are computed from full-precision values; displayed values are rounded.}
\label{tab:perplexity}
\small
\begin{tabular}{@{}llccc@{}}
\toprule
\textbf{Model} & \textbf{Condition} & \textbf{Loss} & \textbf{Perplexity} & \textbf{$\times$ Baseline} \\
\midrule
\multirow{3}{*}{\shortstack[l]{Qwen3.5\\0.8B}}
  & Baseline        & 2.03  & 7.6       & 1.0$\times$       \\
  & Linear off      & 12.50 & 268{,}337 & 35{,}200$\times$ \\
  & Attention off   & 6.44  & 625       & 82$\times$       \\
\midrule
\multirow{3}{*}{\shortstack[l]{Falcon-H1\\0.5B}}
  & Baseline        & 1.73  & 5.6       & 1.0$\times$        \\
  & SSM off         & 5.69  & 295       & 53$\times$        \\
  & Attention off   & 2.88  & 17.7      & 3.2$\times$       \\
\bottomrule
\end{tabular}
\end{table}

\begin{figure}[!htbp]
    \centering
    \includegraphics[width=0.75\textwidth]{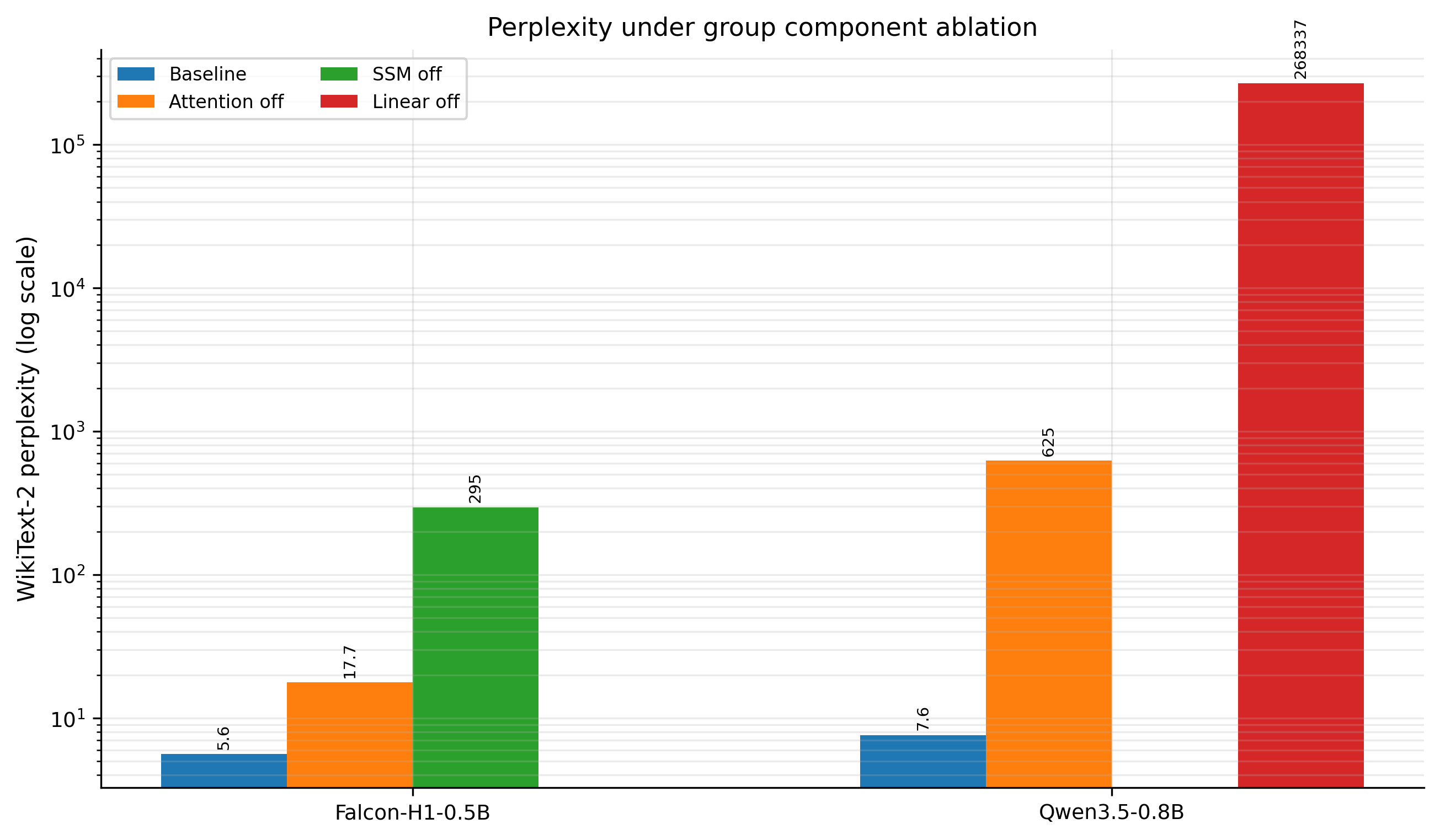}
    \caption{WikiText-2 perplexity under group-level component ablation, shown on a logarithmic scale. Removing the linear-attention pathway in Qwen3.5-0.8B and the SSM pathway in Falcon-H1-0.5B produces larger likelihood degradation than removing attention in the corresponding model. Multipliers are computed from full-precision values; displayed table values are rounded.}
    \label{fig:perplexity}
\end{figure}

For Qwen, removing linear attention increases perplexity from 7.6 to 268{,}337---a factor of approximately 35{,}200$\times$---while removing attention increases it approximately 82$\times$ to 625. For Falcon, removing SSM increases perplexity approximately 53$\times$ to 295, while removing attention increases it only 3.2$\times$ to 17.7. Figure~\ref{fig:perplexity} visualizes these differences on a logarithmic scale.

These results indicate that the alternative sequence-processing pathway is the dominant contributor to next-token likelihood under the tested interventions. Attention remains functionally important for downstream tasks, but its removal causes substantially smaller likelihood degradation than removal of the linear-attention or SSM pathway. This asymmetry is consistent with the findings of Wang et al.~\cite{wang2025systematic}, who showed that recall performance in hybrids improves significantly with even a small proportion of full attention layers, suggesting that attention can provide targeted high-leverage contributions without serving as the dominant likelihood pathway.

\FloatBarrier
\subsection{Matched Random Controls Separate Layer Count from Component Identity}
\label{sec:random_controls}

To separate component-specific effects from the generic effect of disrupting a comparable amount of computation, we compare group ablations against matched random controls (Table~\ref{tab:random_controls}).

\begin{table}[!htbp]
\centering
\caption{Comparison of targeted ablation versus matched random controls (mean $\pm$ standard deviation over 3 trials, averaged across all 5 benchmarks). Random controls remove the same number of computational units as the corresponding targeted condition. The comparison separates degradation due to the quantity and position of disrupted units from degradation associated with a specific component type.}
\label{tab:random_controls}
\small
\begin{tabular}{@{}llcc@{}}
\toprule
\textbf{Model} & \textbf{Condition} & \textbf{Targeted Mean} & \textbf{Random Ctrl Mean} \\
\midrule
\multirow{2}{*}{\shortstack[l]{Qwen3.5\\0.8B}}
  & Linear off (18 layers)    & .202  & .199 $\pm$ .011 \\
  & Attention off (6 layers)  & .210  & .206 $\pm$ .005 \\
\midrule
\shortstack[l]{Falcon-H1\\0.5B}
  & SSM off (all blocks)      & .198  & .203 $\pm$ .012 \\
\bottomrule
\end{tabular}
\end{table}

\begin{figure}[!htbp]
    \centering
    \includegraphics[width=\textwidth]{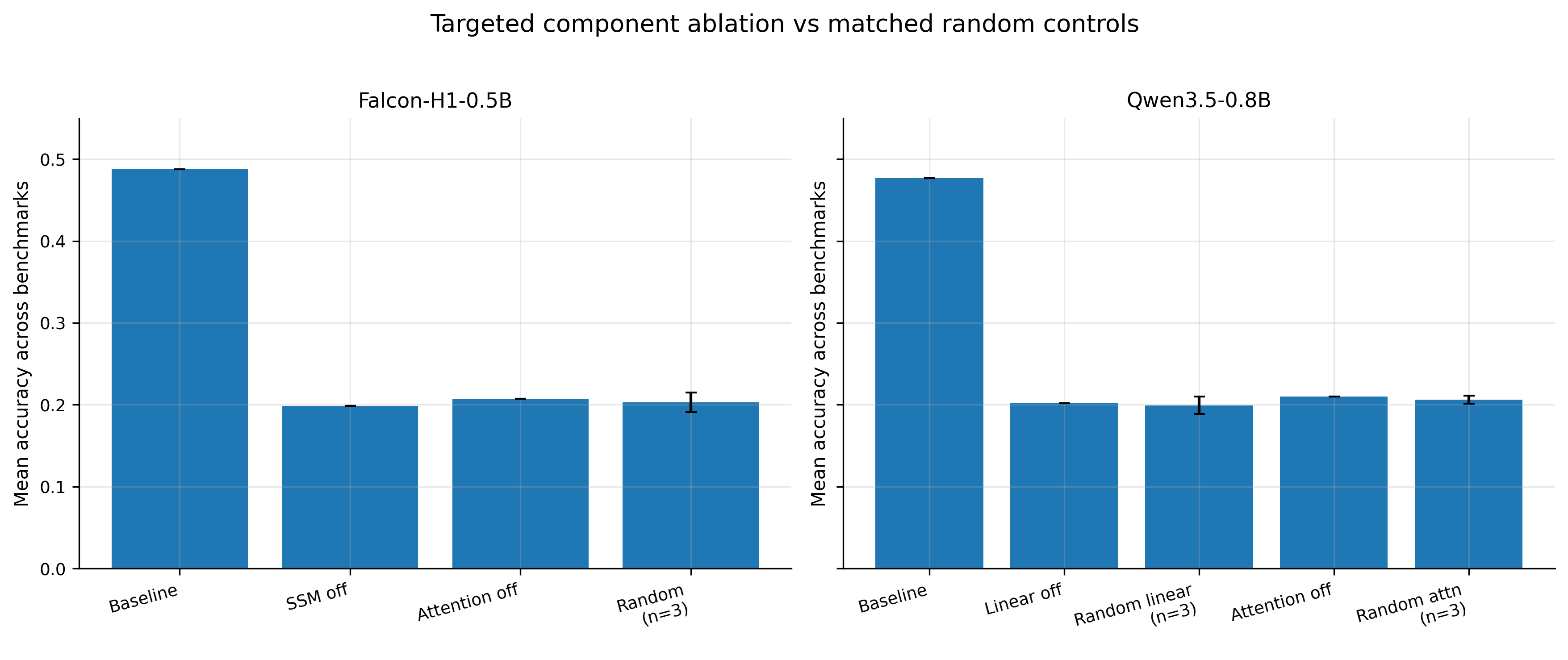}
    \caption{Mean benchmark accuracy under targeted component ablation and matched random ablation. Random controls remove the same number of computational units as the corresponding targeted condition and are averaged over trials; error bars show standard deviation across random trials. The comparison separates degradation due to the quantity and position of disrupted units from degradation associated with a specific component type.}
    \label{fig:random_controls}
\end{figure}

The random controls yield similar mean accuracy to the targeted ablations across all conditions. For Qwen, removing 18 random layers produces a mean accuracy of .199 $\pm$ .011, nearly identical to the .202 from targeted linear-layer removal. This indicates that aggregate benchmark accuracy is strongly affected by the amount and position of disrupted computation, not only by functional type. Figure~\ref{fig:random_controls} illustrates this comparison visually.

However, \textit{perplexity} tells a starkly different story. Table~\ref{tab:rc_perplexity} presents the perplexity under matched random controls for Qwen (5 trials each). Crucially, removing 18 random layers causes \textit{far greater} perplexity degradation (mean 1,748,920) than removing the 18 targeted linear layers (268,337)---a 6.5$\times$ difference. Similarly, removing 6 random layers yields a mean perplexity of 41,154, compared to only 625 for targeted attention removal---a 66$\times$ difference. This suggests that targeted component removal can be less damaging than arbitrary layer removal of the same count, but the high variance of the 6-layer random controls requires cautious interpretation. In mean PPL, attention removal is associated with a larger random-control gap (approximately 66$\times$) than linear-layer removal (approximately 6.5$\times$).

\begin{table}[!htbp]
\centering
\caption{WikiText-2 perplexity under targeted vs.\ matched random ablation for Qwen3.5-0.8B (mean $\pm$ standard deviation over 5 trials). Targeted removal causes less mean PPL damage than random removal of the same number of layers. Ratios are computed from mean PPL values; the 6-layer random control has high variance because some samples include critical early layers.}
\label{tab:rc_perplexity}
\small
\begin{tabular}{@{}lccc@{}}
\toprule
\textbf{Condition} & \textbf{Targeted PPL} & \textbf{Random PPL} & \textbf{Redundancy} \\
\midrule
Linear off (18 layers)    & 268{,}337 & 1{,}748{,}920 $\pm$ 766{,}270 & 6.5$\times$ \\
Attention off (6 layers)  & 625       & 41{,}154 $\pm$ 81{,}609       & 66$\times$  \\
\bottomrule
\end{tabular}
\end{table}

The high variance in the 6-layer random controls is informative: trial~0, which happened to include layer~0 (a critical early linear layer), produced a perplexity of 204,370, while trials excluding layer~0 averaged only 350. This supports the positional importance gradient identified in Section~\ref{sec:layer_sweep}.

\paragraph{Transformer baseline} To determine whether the observed redundancy patterns are specific to hybrid architectures or general properties of deep networks, we repeat the random layer removal experiment on a pure Transformer of comparable size: Qwen2.5-0.5B~\cite{qwen2025qwen25} (490M parameters, 24 layers, softmax attention only). Table~\ref{tab:transformer_baseline} presents the results.

\begin{table}[!htbp]
\centering
\caption{Perplexity under random layer removal: hybrid (Qwen3.5-0.8B) vs.\ pure Transformer (Qwen2.5-0.5B). Values are normalized as multiples of each model's baseline perplexity to enable cross-architecture comparison. Under this protocol, the Transformer control shows 20--119$\times$ larger normalized PPL increases than the hybrid model.}
\label{tab:transformer_baseline}
\small
\begin{tabular}{@{}lccc@{}}
\toprule
\textbf{Layers removed} & \textbf{Hybrid ($\times$ base)} & \textbf{Transformer ($\times$ base)} & \textbf{Ratio} \\
\midrule
18 of 24 & 229{,}408$\times$ & 4{,}701{,}068$\times$ & 20$\times$ \\
6 of 24  & 5{,}398$\times$   & 640{,}850$\times$     & 119$\times$ \\
\bottomrule
\end{tabular}
\end{table}

The pure Transformer control is more sensitive under this protocol: removing 6 random layers degrades it 640,763$\times$ (vs.\ 5,401$\times$ for the hybrid), a 119-fold difference. This suggests greater normalized tolerance to random layer removal in the tested hybrid model under this protocol. A plausible interpretation is that the surviving component type in neighboring layers partially compensates for structural disruption, but this mechanism should be validated in additional model families.

\FloatBarrier
\subsection{Layer-wise Analysis: Positional Sensitivity}
\label{sec:layer_sweep}

\begin{figure}[!htbp]
    \centering
    \includegraphics[width=\textwidth]{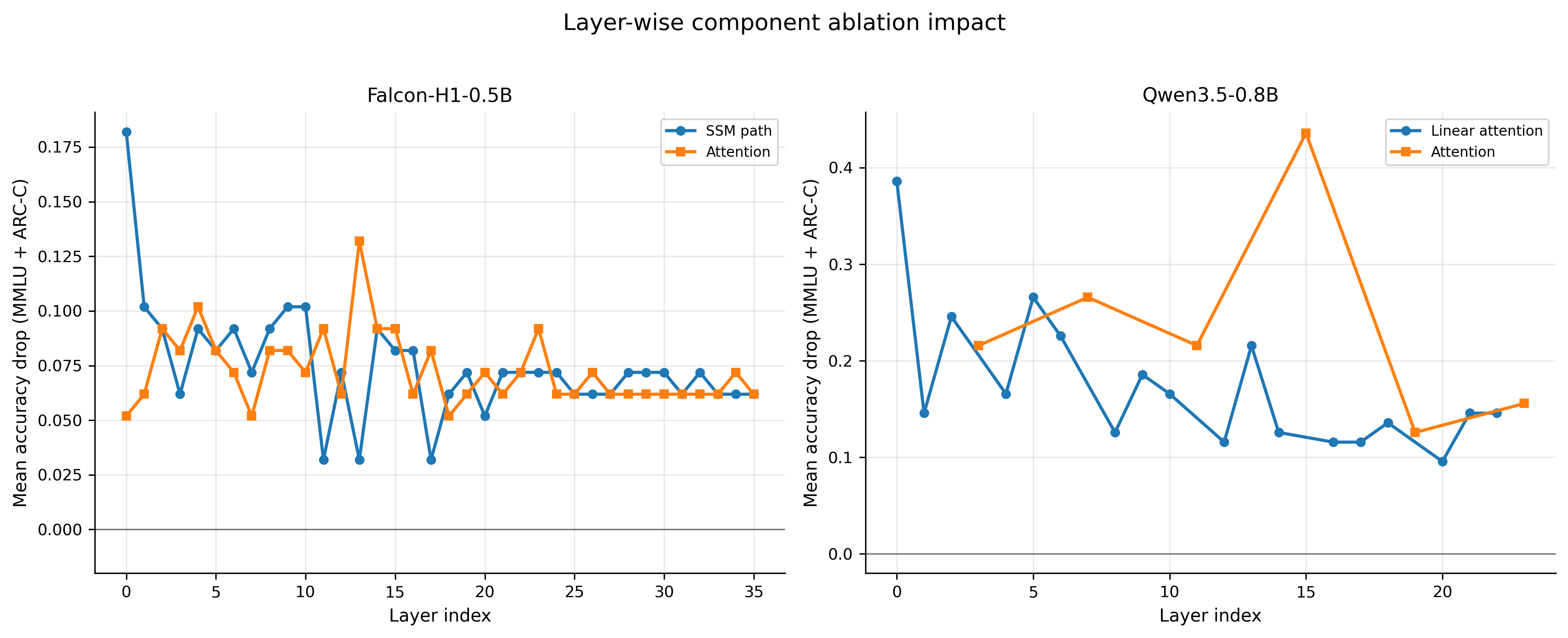}
    \caption{Layer-wise ablation impact, measured as the mean accuracy drop from baseline across MMLU and ARC-Challenge after ablating one component at one layer. Points correspond only to layers where the component is present. Larger positive values indicate greater degradation. The strongest effects are concentrated in early or mid-network components, while late layers generally show smaller drops.}
    \label{fig:layer_sweep}
\end{figure}

Figure~\ref{fig:layer_sweep} presents the accuracy drop from ablating a single component in a single layer, sweeping across all layers of both models. The main pattern is positional sensitivity: many of the largest drops are concentrated in early or mid-network components, while late components generally produce smaller average degradation.

For Qwen, single-layer sensitivity is concentrated in a small subset of early and mid-depth layers. Attention layer~15 produces the largest average drop across MMLU and ARC-Challenge, followed by linear layer~0. Attention layers~7 and~3, as well as several early linear layers, also show elevated sensitivity. This indicates that the sequential hybrid architecture does not distribute importance uniformly across depth: a small number of linear and attention layers are disproportionately important under single-layer removal.

For Falcon, the first SSM layer produces the largest single-layer degradation, and SSM ablations show a clearer early-layer sensitivity pattern than attention ablations. Attention ablations are more evenly distributed, with several mid-depth attention layers also showing measurable impact. Overall, Falcon's parallel hybrid design appears to distribute functional importance more broadly across components than Qwen's sequential design, although early SSM layers remain especially important.

\begin{figure}[!htbp]
    \centering
    \includegraphics[width=\textwidth]{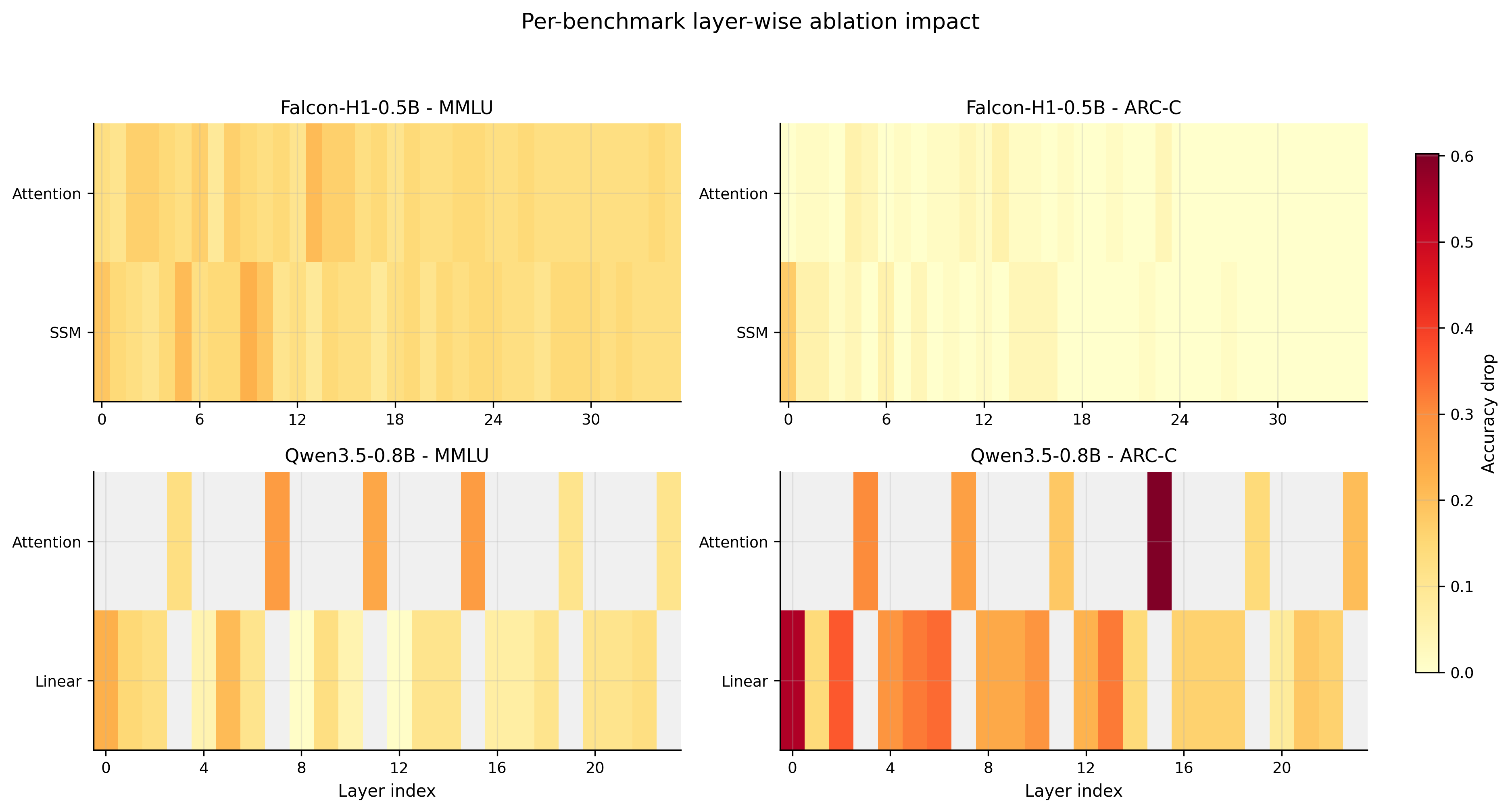}
    \caption{Per-benchmark layer-wise ablation impact for each model. Cell values show score drop from the corresponding baseline after ablating a single component at the indicated layer. Warmer colors indicate larger degradation; blank cells indicate layers where that component type is not present. The heatmaps provide a task-specific view of the positional sensitivity summarized in Figure~\ref{fig:layer_sweep}.}
    \label{fig:heatmap}
\end{figure}

The detailed heatmap in Figure~\ref{fig:heatmap} provides a per-benchmark view of this layer-wise sensitivity across both architectures.

\FloatBarrier
\subsection{Positional Ablations (Falcon)}
\label{sec:positional}

Table~\ref{tab:positional} presents the positional ablation results for Falcon, where all components of one type within the early, middle, or late third of layers are simultaneously removed.

\begin{table}[!htbp]
\centering
\caption{Falcon-H1 positional ablation: removing all components of one type within a positional third. Early-layer ablation causes substantially more damage than late-layer ablation for both SSM and attention.}
\label{tab:positional}
\small
\begin{tabular}{@{}llccccc@{}}
\toprule
\textbf{Component} & \textbf{Position} & \textbf{MMLU} & \textbf{GSM8K} & \textbf{ARC-C} & \textbf{HSwag} & \textbf{TFQA} \\
\midrule
\multicolumn{2}{l}{\textit{Baseline}} & .568 & .400 & .736 & .512 & .223 \\
\midrule
\multirow{3}{*}{SSM off}
  & Early  & .255 & .020 & .294 & .274 & .248 \\
  & Middle & .454 & .280 & .592 & .457 & .224 \\
  & Late   & .538 & .160 & .676 & .485 & .234 \\
\midrule
\multirow{3}{*}{Attn off}
  & Early  & .278 & .040 & .244 & .442 & .230 \\
  & Middle & .441 & .000 & .555 & .412 & .225 \\
  & Late   & .553 & .260 & .659 & .492 & .235 \\
\bottomrule
\end{tabular}
\end{table}

The results support the layer-sweep findings quantitatively: removing early SSM reduces MMLU from .568 to .255 ($-$55\%), while removing late SSM only reduces it to .538 ($-$5\%). The asymmetry is even more dramatic for ARC-Challenge: early SSM off yields .294 ($-$60\%) versus late SSM off at .676 ($-$8\%). Attention shows a similar gradient but with a notable exception: middle attention ablation causes GSM8K to collapse to 0.000, suggesting that middle-layer attention is specifically critical for mathematical reasoning chains.

\FloatBarrier
\subsection{Hidden-State Diagnostics}
\label{sec:exp2}

To complement behavioral ablation with internal analysis, we compute layer-wise hidden-state metrics using a diagnostic corpus (512 tokens from WikiText-2). For each layer, we measure four quantities:
\begin{align}
\text{norm change} &= \|\mathbf{h}_\text{post} - \mathbf{h}_\text{pre}\| \,/\, \|\mathbf{h}_\text{pre}\| \nonumber \\
\text{cosine sim.} &= \cos(\mathbf{h}_\text{pre}, \mathbf{h}_\text{post}) \nonumber \\
\text{output ratio} &= \|f_c(\mathbf{h})\| \,/\, \|\mathbf{h}\| \quad \text{for each component } c \nonumber \\
\text{logit-lens KL} &= \mathrm{KL}(p_\text{baseline} \| p_\text{ablated}) \nonumber
\end{align}

For Qwen, the first linear layer (layer~0) stands out with a norm change of 1.98, far exceeding all other layers. Excluding this outlier, linear and attention layers show similar mean norm change (0.50 vs.\ 0.59), but the component output norm ratio is higher for linear layers (mean 0.44 vs.\ 0.36 for attention), indicating that linear layers contribute proportionally more to the residual stream per layer.

For Falcon, the SSM path shows dramatically higher logit-lens KL in late layers (layer~30: KL=32.0; layer~33: KL=11.1), indicating that the SSM contributes most to next-token prediction in the model's final processing stages. The attention path shows its highest KL at layer~0 (KL=0.71), suggesting that early attention plays a distinct role in initial token contextualization.

\begin{figure}[!htbp]
    \centering
    \includegraphics[width=\textwidth]{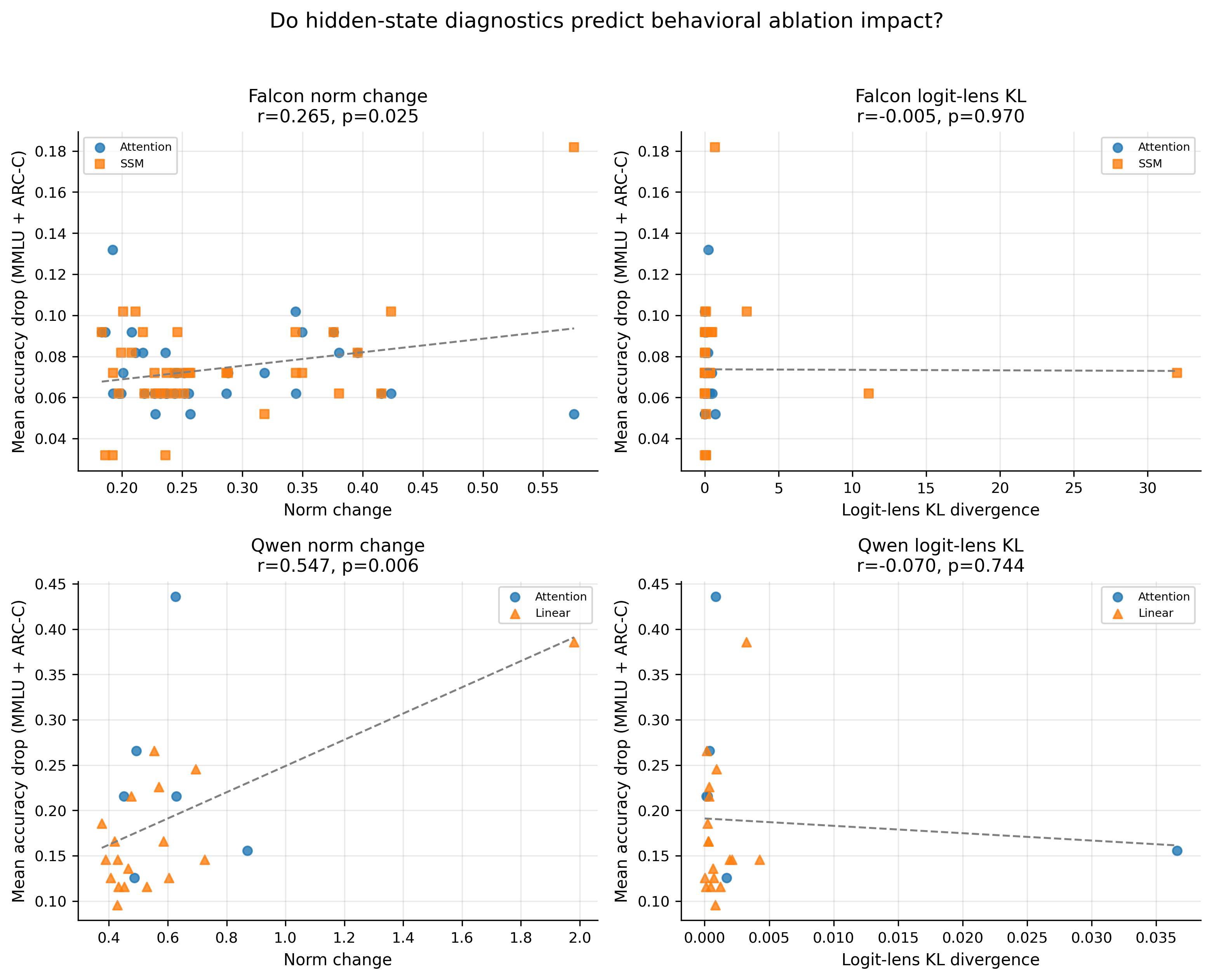}
    \caption{Relationship between layer-wise hidden-state diagnostics and behavioral ablation impact. Each point corresponds to one ablated component/layer condition. The $x$-axis reports either normalized hidden-state change or logit-lens KL divergence; the $y$-axis reports mean accuracy drop across MMLU and ARC-Challenge. Pearson correlations and $p$-values are computed over the plotted points. Norm change correlates positively with ablation impact in Qwen and more weakly in Falcon, whereas logit-lens KL does not significantly predict ablation impact in either model.}
    \label{fig:correlation}
\end{figure}

Figure~\ref{fig:correlation} examines the relationship between these internal metrics and the behavioral ablation results. For Qwen, norm change shows a significant positive correlation with accuracy drop ($r=0.547$, $p=0.006$), indicating that layers that transform representations more strongly also tend to cause larger downstream degradation when ablated. Logit-lens KL divergence does not significantly predict ablation impact in Qwen ($r=-0.070$, $p=0.744$).

For Falcon, norm change is measured at the hybrid-block level and is therefore matched to component ablations at the corresponding layer. Under this descriptive analysis, block-level norm change shows a weaker but significant positive correlation with ablation impact ($r=0.265$, $p=0.025$). In contrast, component-level logit-lens KL divergence is not predictive of behavioral degradation ($r=-0.005$, $p=0.970$). These results suggest that hidden-state magnitude diagnostics can provide partial information about layer sensitivity, but likelihood-oriented logit-lens measures do not reliably identify which component ablations will harm benchmark performance.

\FloatBarrier
\subsection{Task-Dependent Component Sensitivity}
\label{sec:exp3}

\begin{figure}[!htbp]
    \centering
    \includegraphics[width=0.85\textwidth]{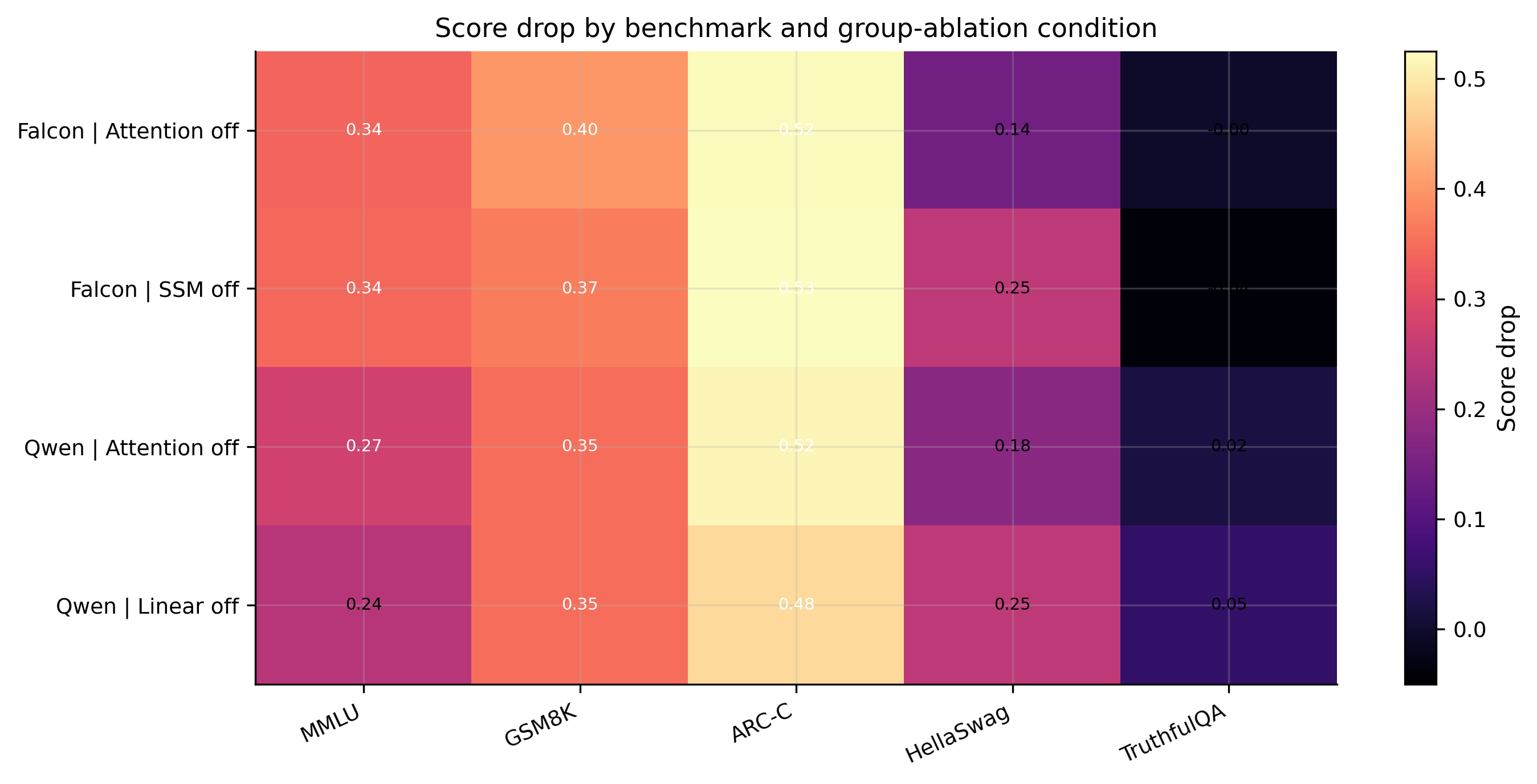}
    \caption{Score drop by benchmark and group-ablation condition. Columns correspond to benchmarks and rows to model--condition pairs. Warmer colors indicate larger degradation relative to the corresponding baseline; negative values indicate improvement under ablation. In the tested models, GSM8K shows the largest drops, HellaSwag shows clear component asymmetry, and TruthfulQA is comparatively insensitive to several ablations.}
    \label{fig:exp3_heatmap}
\end{figure}

We analyze whether specific components are differentially important for specific task types. Figure~\ref{fig:exp3_heatmap} shows the score drop heatmap across all model-condition-benchmark combinations. Across all conditions, we observe three patterns:

\textbf{Mathematical reasoning (GSM8K)} is the most fragile capability in these evaluations: its score falls to 0\% under nearly all group ablations in both models, indicating high sensitivity to structural removal on the evaluated sample.

\textbf{Commonsense reasoning (HellaSwag)} shows the most asymmetric response: removing attention has a substantially smaller effect than removing the alternative component in both models (Qwen: $-$.175 vs.\ $-$.248; Falcon: $-$.143 vs.\ $-$.247), suggesting greater dependence on the recurrent/linear pathway under these interventions.

\textbf{Factual discrimination (TruthfulQA)} is the most robust to ablation, with drops of $<$6\% in most conditions and even slight \textit{improvements} in Falcon under SSM-off (+4.1\%), possibly reflecting reduced model confidence in misleading completions.

\FloatBarrier
\section{Discussion}
\label{sec:discussion}

\paragraph{Both component types are functionally used}
In the tested natively trained hybrids, removing either attention or the alternative sequence-processing pathway substantially degrades performance. This contrasts with observations in some post-training linearized models~\cite{benfeghoul2025untangling}, where the linear component can be weakly used or bypassed. Our results suggest that native hybrid training can produce measurable functional reliance on both component types, although the degree of reliance varies across metrics and tasks.

\paragraph{Likelihood sensitivity is asymmetric}
The largest asymmetry appears in WikiText-2 perplexity. Removing Qwen's linear-attention layers and Falcon's SSM pathway produces substantially larger likelihood degradation than removing attention alone. We therefore describe the linear-attention/SSM pathway as the dominant contributor to next-token likelihood under the tested interventions, rather than as a universally primary component. Attention remains important for downstream performance, and its smaller perplexity effect is compatible with a complementary, high-leverage role.

\paragraph{Implications for compression and efficient deployment}
The positional gradient in component importance suggests that pruning strategies for hybrid models should be position-aware. In the tested models, early-layer components are more sensitive to removal than later components. This implies that compression, quantization, and distillation methods should not treat all component types or layer positions as interchangeable. The results provide empirical constraints for deciding which components require higher fidelity and which may be candidates for more aggressive compression.

\paragraph{Sequential versus parallel hybridization}
The two architectures expose different intervention units and show different degradation profiles. Qwen's sequential design creates distinct component positions, enabling direct layer-type comparisons but also coupling each token mixer with its layer-local feed-forward transformation. Falcon's parallel design allows path-specific zeroing within each block, preserving the other path and the block-level feed-forward computation. These differences mean that cross-model magnitudes should be interpreted cautiously. The more robust conclusion is that both hybridization strategies exhibit component-specific and position-specific degradation patterns.

\paragraph{Benchmark accuracy and perplexity measure different effects}
A methodological insight from the study is that standard benchmarks and perplexity reveal complementary aspects of component function. Benchmark accuracy measures prompted task behavior, whereas perplexity measures next-token likelihood under a standardized language-modeling corpus. Similar benchmark drops can coexist with very different likelihood degradation. Future component-analysis studies should therefore include both task-level and likelihood-level diagnostics.

\paragraph{Hybrid tolerance under random structural removal}
The Transformer baseline comparison suggests that the tested hybrid model has greater normalized tolerance to matched random layer removal than the same-family Transformer control. We interpret this as evidence that hybridization can alter degradation behavior under structural perturbation. The result should not be read as a universal fault-tolerance guarantee, but it motivates further study of graceful degradation and recovery training in efficient hybrid architectures.

Overall, the results support a component-aware view of hybrid language model deployment. In the tested models, early or mid-network components and the linear-attention/SSM pathways are most sensitive under ablation, suggesting that compression and pruning should not treat all layers or component types as interchangeable. The Transformer control further indicates that hybridization can change degradation behavior under random structural removal. These observations do not by themselves define an optimal compressed architecture, but they provide empirical constraints for pruning, distillation, and fault-tolerance studies in efficient hybrid language models.

\FloatBarrier
\section{Limitations and Future Work}
\label{sec:limitations}

\paragraph{Scale and model-family scope} This study uses sub-1B models to make exhaustive component ablation feasible and reproducible. The results identify consistent patterns in two representative hybrid designs, but they should not be interpreted as proving that the same quantitative ratios hold at 3B, 7B, or larger scales. Larger models may redistribute component importance with depth, training data, or hybridization ratio. Future work should validate the most important ablation conditions at larger scale.

\paragraph{Functional ablation rather than retraining} Our interventions measure inference-time functional dependence. Removing or zeroing a component after training is not equivalent to training an architecture without that component. The reported degradation therefore estimates sensitivity to structural removal, not the best achievable performance of a retrained smaller model. This distinction is central for interpreting the results as compression guidance: ablation identifies which components are risky to remove, while pruning or distillation would require recovery training.

\paragraph{Architecture-specific interventions} The two architectures expose different intervention units. Qwen's sequential skip removes a full layer transformation, whereas Falcon's parallel zeroing removes a component path while preserving the rest of the block. We therefore emphasize within-model comparisons and matched random controls rather than direct equality of ablation magnitudes across architectures.

\paragraph{Benchmark and sample coverage} The benchmark suite covers knowledge, mathematical reasoning, science reasoning, commonsense completion, and factual discrimination, but it does not exhaust all deployment workloads. Some GSM8K and positional conditions use smaller samples because of the large number of ablation conditions. These results should be interpreted as diagnostic evidence of relative degradation rather than final task-performance estimates.

\paragraph{Context length} The evaluated prompts are short relative to the maximum context lengths supported by the models. This limits conclusions about long-context behavior, where linear-attention and SSM pathways may play different roles. Future work should evaluate matched ablations across longer contexts and retrieval-style tasks.

\paragraph{No recovery after ablation} We do not fine-tune, distill, or otherwise adapt the model after removing components. This choice isolates immediate functional dependence, but it does not measure how much performance could be recovered after structured pruning. A natural next step is to combine the present diagnostics with recovery fine-tuning.

\paragraph{Base-model setting} We analyze base checkpoints rather than instruction-tuned or RLHF-tuned variants. Instruction tuning may alter task sensitivity and component reliance. Future work should test whether the same qualitative patterns persist after alignment and deployment-oriented fine-tuning.

\FloatBarrier
\section{Conclusion}
\label{sec:conclusion}

This work presents a component-level ablation analysis of two hybrid language model architectures: Qwen3.5-0.8B, a sequential hybrid with Gated DeltaNet and softmax attention layers, and Falcon-H1-0.5B, a parallel hybrid with Mamba-2 and attention paths. Using group-level, layer-wise, positional, and matched random ablations, together with a same-family Transformer control, we evaluate how structural component removal affects downstream accuracy, WikiText-2 perplexity, and hidden-state diagnostics.

The results show that both component types are functionally used in the tested natively trained hybrids. Removing either attention or the alternative sequence-processing pathway substantially degrades downstream performance. At the likelihood level, however, the degradation is asymmetric: removing the linear-attention/SSM pathway produces larger WikiText-2 perplexity increases than removing attention alone, suggesting that these pathways dominate next-token likelihood under the tested interventions. Layer-wise and positional analyses further show that many of the largest ablation effects are concentrated in early or mid-network components, providing empirical constraints for position-aware compression and distillation. Finally, matched random controls and a Transformer baseline indicate that architecture type affects degradation under structural removal.

These findings support a component-aware view of efficient hybrid language model deployment. Compression and pruning strategies should account for component type, layer position, and metric-specific sensitivity rather than treating all layers as interchangeable. More broadly, functional ablation provides a practical diagnostic for studying resilience, specialization, and compression implications in emerging hybrid neural architectures.

\FloatBarrier
\section*{Data and Code Availability}

Code used to implement reversible ablations, evaluation scripts, configuration files, random seeds, and aggregate results supporting the tables and figures are available at:
\url{https://github.com/hecboar/hybrid-component-ablation}.
The repository does not redistribute pretrained model weights or benchmark datasets; these are accessed from their original providers under their respective licenses. Additional execution logs and large intermediate artifacts are available from the corresponding author upon reasonable request due to storage requirements.

\section*{Declaration on the Use of Generative AI}
During manuscript preparation, generative AI tools were used to assist with language editing, structural refinement, and consistency checking. The authors reviewed and edited all AI-assisted output, verified the reported results and references, and take full responsibility for the final manuscript. No generative AI tool was used as an author, to generate research data, or to create figures.

\section*{Competing Interests}
The authors declare that they have no competing financial or non-financial interests.

\section*{Funding}
The authors received no specific funding for this work.

\section*{Author Contributions}
Hector Borobia: conceptualization, methodology, software, validation, formal analysis, investigation, data curation, writing--original draft, writing--review and editing, visualization. Elies Segu\'i-Mas: supervision, writing--review and editing. Guillermina Tormo-Carb\'o: supervision, writing--review and editing.

\section*{Acknowledgements}
The authors thank the Universitat Polit\`ecnica de Val\`encia for institutional support.


\bibliographystyle{elsarticle-num}

\end{document}